# Cluster-Based Information Retrieval by using (K-means)-Hierarchical Parallel Genetic Algorithms Approach

**Sarah Hussein Toman[1], Mohammed Hamzah Abed[2], Zinah Hussein Toman[3],**

[1] Roads and Transport Department, College of Engineering, University of Al –Qadisiyah, Ad-Diwaniah, Iraq
[2,3] Computer Science Department, College of Computer Science and Information Technology,
University of Al –Qadisiyah/Al-Diwaniah, Iraq

**ABSTRACT**

Cluster-based information retrieval is one of the Information retrieval(IR) tools that organize, extract features and categorize the web documents according to their similarity. Unlike traditional approaches, cluster-based IR is fast in processing large datasets of document. To improve the quality of retrieved documents, increase the efficiency of IR and reduce irrelevant documents from user search. in this paper, we proposed a (K-means) - Hierarchical Parallel Genetic Algorithms Approach (HPGA) that combines the K-means clustering algorithm with hybrid PG of multi-deme and master/slave PG algorithms. K-means uses to cluster the population to k subpopulations then take most clusters relevant to the query to manipulate in a parallel way by the two levels of genetic parallelism, thus, irrelevant documents will not be included in subpopulations, as a way to improve the quality of results. Three common datasets (NLP, CISI, and CACM) are used to compute the recall, precision, and F-measure averages. Finally, we compared the precision values of three datasets with Genetic-IR and classic-IR. The proposed approach precision improvements with IR-GA were 45% in the CACM, 27% in the CISI, and 25% in the NLP. While, by comparing with Classic-IR, (k-means)-HPGA got 47% in CACM, 28% in CISI, and 34% in NLP.

*Corresponding Author:*

Sarah Hussein Toman,
Roads and Transport Department,
College of Engineering, University of Al –Qadisiyah,
Ad-Diwaniah, Iraq.
Email: sarah.toman@qu.edu.iq

## 1. INTRODUCTION

In the recent years, the information has been overloaded because of the rapid growth of the web. To deal with this information a Web Document Information Retrieval task is used to retrieve the most relevant documents to a user query[1,2]. Information retrieval needs to scan all documents that are found in a database, then give scores according to a relevance degree to the user query, then rank all results and present them to the user[3,4]. Thus, information retrieval requires long runtime to scan all documents. The cluster analysis tool plays a basic role in information retrieval to improve the Information Retrieval performance by reducing the search time and to prevent irrelevant results from the retrieved documents. The idea behind the web document clustering is to assign a dataset of web documents to a set of clusters that depend on the similarity's degree among them. Therefore, it becomes easy for search engines to query in the same cluster if each web page is assigned to a similar group[5,6].

An efficient clustering algorithm and genetic algorithm should represent a document as structured data using the document representation model. The most common aspect used in document representation is the Vector Space Model (VSM)[7]. Besides, a similarity degree between two documents or clusters should be measured by using one of the similarity measures[1].

Hierarchical and partition algorithms are the major kinds of clustering algorithms have been used[8]. A hierarchical clustering algorithm generates a tree of clusters (groups) depending on two methods. The first method starts with one cluster then merges each two similar clusters, which is known as the agglomerative method. The second one starts from the whole data set as one cluster then split it into clusters at each stage, is known as the divisive method [9,10]. A partition clustering algorithm uses a single step to divide the collection of documents in to predefined number of groups[11]. The most widely used partition clustering algorithm is the K-means algorithm[12]. It is an unsupervised learning algorithm that relies on selecting K clusters as K-centroids. After that, the similarity measure is calculated between each document and the centroids, then the documents will assign to the closest centroid after updating of centroids multiple times [13].

In the present paper, the k-means cluster with two levels of genetic parallel is used for information retrieval. Multi-deme parallel genetic as first level and Master-Slave parallel genetic as second level. The idea behind using the K-mean clustering algorithm is to group a set of documents to clusters according to their similarity with a query, then an HPGA algorithm will perform a search in the most relevant clusters to reduce the search time and to provide optimal search results. Next, at each subpopulation there is a fitness evaluation parallelism with hybrid selection and two chromosomes crossover as genetic operators. Then migration among individuals and repeat HPGA steps n time until obtaining the optimal results.

## 2. TERM FREQUENCY – INVERSE DOCUMENT FREQUENCY (TF-IDF)

Datasets in most clustering algorithms are represented by a set of vectors, $V = \{V1, V2, V3… Vn\}$, where, Vi is the feature vector of one object. Term Frequency is a simple and effective term selection method, alike words are used in the documents that belong to the same subject, thus, term frequency can be a respectable indicator for a certain subject. TF is a term occurrence frequency in the document as shown in equation 1. On another hand, some terms should be removed such as words in the stop list corresponding to the English language, because the occurrence of these words is not relevant to identify the subject of the document[14].

$$TF(j, i) = \text{frequency of } i\text{ th term in document } j \tag{1}$$

TF is not effective to measure the frequent terms in a set of documents. Thus, Inverse Document Frequency (IDF) is used. TDF is the term frequency across a set of documents as shown in equation (2).

$$IDF(t_i) = \log \frac{|D|}{|D_{t_i}|} \tag{2}$$

*|D|, number of documents.*
*|Dti|, number of documents that contain the term ti.*

To determine the weight for each term ti in each document dj, TF and IDF will be combined by multiplication of the resulted values, TF-IDF given by equation 3[15]. In document clustering, terms with higher TD-IDF have better clustering.

$$TF\text{-}IDF(t_i, d_j) = TF(j, i) * IDF(t_i) \tag{3}$$

## 3. GENETIC ALGORITHM

The genetic algorithm (GA) is a probabilistic meta-heuristic search algorithm inspired by natural genetics[16,17]. GA gives a good solution in many life fields. Figure 1 demonstrates the flowchart of the genetic algorithm steps. The basic operations of a genetic algorithm are [18,19]:

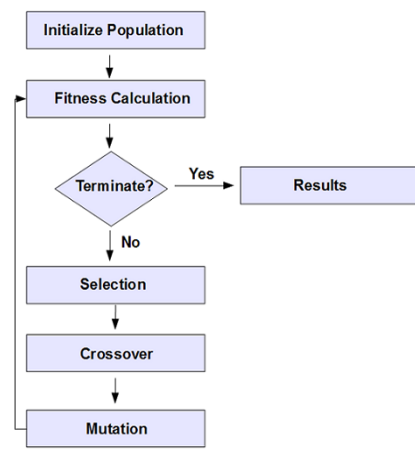

1. Generate random solutions that are called a population.
2. Determine Fitness value to evaluate each solution.
3. Select the best solutions according to the fitness.
4. Produce a new population by genetic operators (crossover and mutation).

As employ the parallelism feature to reduce the process duration. There are three models of Parallel Genetic Algorithms (PGA) as exhibited in figure (2): (a) Master/Slave PGA which deals with single population and parallel fitness calculation; (b) multi deme PGA which deals with multi-population and parallel genetic operations followed by migration among them; (c) Cellular which deals with a single population running on a parallel processing system based closely-linked massively.
The previous models can be hybridized to produce Hierarchical PGA (HPGA) models [20,21].

Figure 1. Genetic algorithm steps

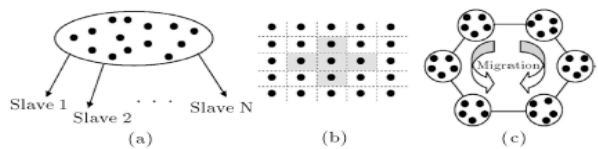

Figure **2.** (a) Master/Slave PGA, (b) Multi deme PGA and (c) Cellular PGA

## 4. THE PROPOSED APPROACH

The Information Retrieval systems process a large amount of text in documents index and user query stages. Parallelism is a way to improve the query average time. The elaborated procedure uses a Parallel Genetic Algorithm with K-means to retrieve the most relevant documents to a user query that relies on the steps enumerated below, Figure 4 presents the proposed (K-mean)-HPGA approach:

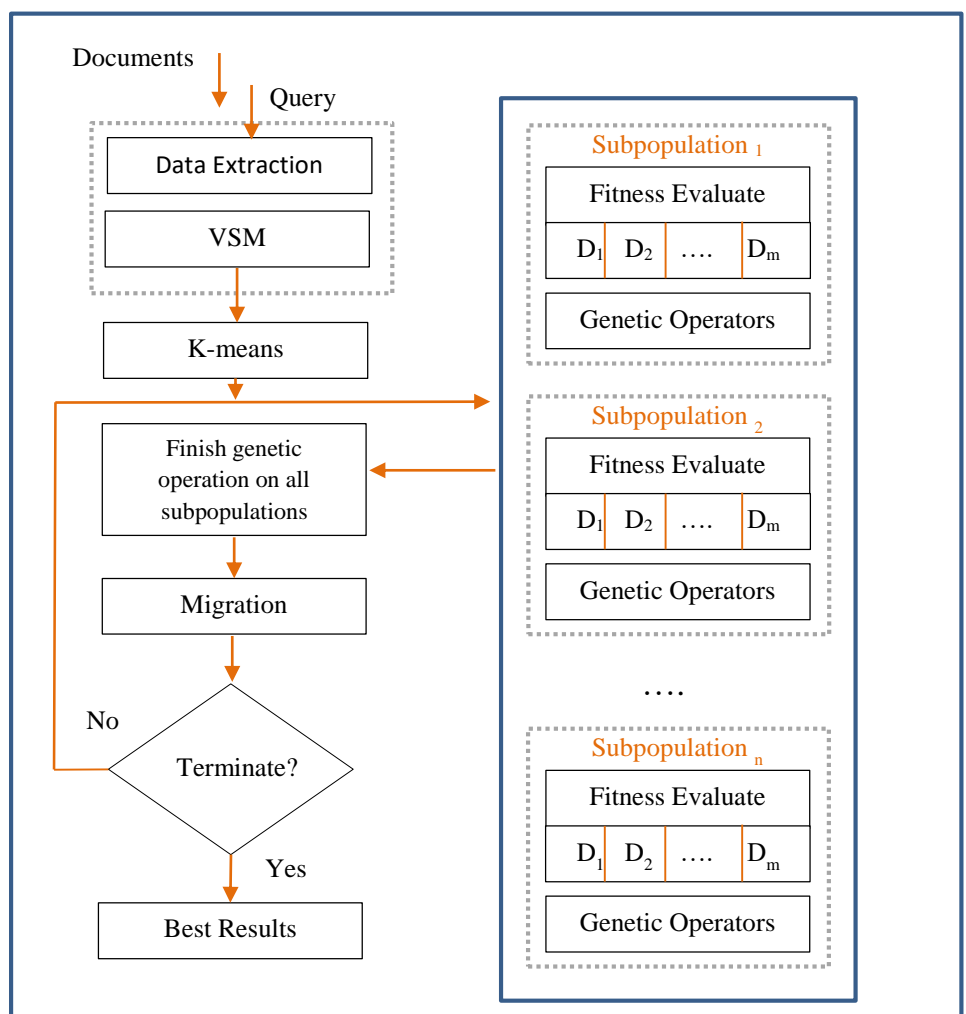

Figure 4. (K-means) - HPGA approach

### 4.1 Web Document Data Extraction

Web page extraction represents the interaction with web page source (HTML) to scrap the information, respectively to identify structured data as a post-processing stage that is composed of two steps:

#### 4.1.1 Tree-based extraction

web pages have a semi-structured feature, therefore, this feature is considered the most important feature to represent the HTML tags and text as a labeled tree, which is called a DOM (Document Object Model)[22], and addressing the element's tag in the tree via XPath language.

#### 4.1.2 Text Tokenizer

its purpose is to break the text in tokens, eliminating stop words and stemmer from tokens. The Stop Wordlist that we used, contains 1300 words which include articles (a, an, the), prepositions (in, into, on, at), conjunctions (and, or, but, and so on), pronouns (she, he, I, me), and other words irrelevant for the query process. Porter Stemming is used in our approach to enhance accuracy via dropping morphological variants of words. Thus, tokens with common stems such as -ED,-ING,-ION, and -IONS will have similar meanings.

### 4.2 Document and Query Representation

In this approach, Vector Space Model (VSM) is used, a features vector is generated from each document content and the given query, depending on the occurrence of words in the document by using TF-IDF function (the frequency occurrence of the term in the document (TF) with the frequency of occurrence of the term in the data set of documents (TF-IDF), as mentioned in equation 3).

### 4.3 K-means - Hierarchical Parallel Genetic Algorithms Approach

The idea behind using the Parallel algorithm is to split the task into a set of subtasks that will exhibit a divide-and-conquer behavior. In our approach we use multi-deme parallel genetic (multiple population) with k-means clustering. Steps bellow explain the algorithm operation in details:

#### 4.3.1 Generate Population

Create the subpopulations from the web document dataset via the K-means algorithm. K-means split the documents to be indexed into k clusters then evaluate the last centroid with a query and select just clusters that are near from the query.

| K-means Algorithm |
|---|
| *Input: $D = \{d_1, d_2, d_3, ..., d_n\}$, set of documents.* <br>       *K: number of clusters.* <br> *Output: $C = \{C_1, C_2, C_3, ..., C_k\}$, set of clusters.* |
| Step1: Let centroid $c_j$ = random number // $j= 1,...,k$ <br> Step2: Foreach ( $d_i$ in D) <br>      Calculate CosDistance($d_i$, $c_j$), $i = 1,..., n, j = 1,...,k$ <br>    end <br> Step3: Assign each document $d_i$ with minCosDistance($d_i$, $c_j$) to cluster $C_j$ <br> Step4: Update centroid $c_j$, for all j <br> Step5: Repeat( step2 and step 3)  Until (no change in cluster Cj) <br> Step6: End. |

#### 3.4.2 Fitness Evaluation

The second level of the Parallel Algorithm is applied to evaluate the fitness function in each cluster (subpopulation), i.e all documents in the cluster will be evaluated at the same time under the slave/master parallel concept. This evaluation starts by forwarding user query to each cluster then calculate the fitness function to each document of the cluster. In the present approach, a cosine similarity function is used as a fitness function[23]. The cosine similarity function is given in equation 4.

$$Scos = \frac{\sum_{i=1}^{n} Pi\, Qi}{\sqrt{\sum_{i=1}^{n} Pi^2} \sqrt{\sum_{i=1}^{n} Qi^2}} \qquad (4)$$

#### 3.4.3 Genetic Operators

generate a new population by applying genetic operators (selection and crossover). To improve genetic performance, we move 4% of chromosomes with the highest probability in the next generation without change (i.e. apply **Elitism Feature**). Genetic Operators in (K-means) – HPGA flow the following steps:.

  a. Calculate the probability for each chromosome, where P[i] = **Fitness**[i] / **Total**
  b. Rank the Probability values and take the top 4% Elitism to avoid the loss of fittest chromosomes in the new population.
  c. Hybrid Roulette - Tournament Selection (**HRTS**): It is the process of selecting a pair of parents from the population to emphasize fitter offsprings in a new population. In our approach we used a hybrid method to take advantage of both selection methods (Roulette wheel and Tournament). The selection process is explained by the following algorithm:

| HRTS Algorithm |
|---|
| *Input: $pop_{size}$, fitness.* <br> *Output: parent1, parent2.* |
| **Begin** <br>  for j = 1 : 2 <br>    r = randi[1, $pop_{size}$] //Select random number for subpopulation <br>    for i = 1 : r <br>        $sum_{fitness}$ = sum (fitness) |

```
        P_sum = randi[1, sum_fitness];
        S = 0; index = 1;
        S = S + fitness[i];   index++;
        if (s < P_sum) goto 10, else subPop[i] = current chromosome
    end
    Parent[j] = maxFitness(subPop) // select parent
  end
End
```

**d. Crossover**

Crossover operation aims to get better offspring by generating a new child from two selected parents. In this approach, we proposed to represent the population as a matrix, each chromosome vector representing a row in the matrix, then select two random positions in the range [1, vector_length]. The crossover is described by the following Algorithm:

**Two Chromosomes Crossover Algorithm**

*Input: subP = subP - Elite Count.*
*Output: offsprings.*

*Begin*
*subP_length = length(subP);*
*repeat*
  *Call selection function to select two parents;*
*10 Call pickTwoPosition (subP_length);*
  *Exchange two positions betweentwo selected parents;*
*until (index <= subPsize) Goto 10;*
*End*
*function [ position1, position2 ] = pickTwoPosition (subP_length)*
*r = randi([1, subP_length],2)// generate 2 random integer numbers to vector r*
*position1 = r(1);*
*position2 = r(2);*
*end*

### 3.4.4 Migration
is the synchronous process that waits for the evaluation of all chromosomes in all subpopulations to exchange the individuals. Migration has an interval that is set to 5 in our approach.

### 3.4.5 Terminate
repeat the previous steps (from fitness to migration) n times, where n is the size of the population, then retrieve the best results after ranking documents according to fitness Probability values.

## 5. EXPERIMENTAL RESULTS

Three datasets were used for experimental results. NPL Dataset(DS1) containing 11,429 electronic engineering documents, CISI Dataset(DS2) with 1,460 computer science documents and CACM dataset(DS3) consisting of 3204 communications documents. To evaluate the web documents retrieval, the Recall, Precision, and F-measure are used for 100 queries in three datasets as defined in the following equations [24,25]:

$$Recall\ (R) = \frac{relevant\ items\ retrieved}{relevant\ items} \quad (5)$$

$$Precision\ (P) = \frac{relevant\ items\ retrieved}{retrieved\ items} \quad (6)$$

$$F\text{-}measure = \frac{2 \cdot R \cdot P}{R+P} \quad (7)$$

The results are shown in tables 1, 2, and 3. For the NPL Dataset(DS1) where precision average is 0.688889 and F-measure average is 2.0667, while in the CISI Dataset(DS2), the precision average was 0.65889 and the F-measure average was 1.97667. Finally, the CACM dataset(DS3) the average for precision and F-measure were 0.748889 and 45.22222 respectively. After the analysis of the previous results, the third dataset gave higher results in both measures.

Table 1. The results of Recall, Precision and F-measure for 100 query in NPL Dataset(DS1) by using (K-means) - HPGA Approach

| Recall | 0.1 | 0.2 | 0.3 | 0.4 | 0.5 | 0.6 | 0.7 | 0.8 | 0.9 | AVG |
|---|---|---|---|---|---|---|---|---|---|---|
| Precision | 0.9 | 0.87 | 0.84 | 0.77 | 0.74 | 0.66 | 0.58 | 0.46 | 0.38 | 0.6888 |
| F-measure% | 2.7 | 2.61 | 2.52 | 2.31 | 2.22 | 1.98 | 1.74 | 1.38 | 1.14 | 2.0666 |

Table 2. The results of Recall, Precision and F-measure for 100 query in CISI Dataset(DS2) by using (K-means)- HPGA Approach

| Recall | 0.1 | 0.2 | 0.3 | 0.4 | 0.5 | 0.6 | 0.7 | 0.8 | 0.9 | AVG |
|---|---|---|---|---|---|---|---|---|---|---|
| Precision | 0.89 | 0.84 | 0.78 | 0.76 | 0.69 | 0.55 | 0.51 | 0.47 | 0.44 | 0.6588 |
| F-measure % | 2.67 | 2.52 | 2.34 | 2.28 | 2.07 | 1.65 | 1.53 | 1.41 | 1.32 | 1.9766 |

Table 3. displays the results of Recall, Precision and F-measure for 100 query in CACM dataset(DS3) by using (K-means) - HPGA Approach

| Recall | 0.1 | 0.2 | 0.3 | 0.4 | 0.5 | 0.6 | 0.7 | 0.8 | 0.9 | AVG |
|---|---|---|---|---|---|---|---|---|---|---|
| Precision | 0.94 | 0.9 | 0.87 | 0.85 | 0.8 | 0.77 | 0.66 | 0.54 | 0.41 | 0.7488 |
| F-measure % | 2.82 | 2.7 | 2.61 | 2.55 | 2.4 | 2.31 | 1.98 | 1.62 | 1.23 | 2.2466 |

We measured the improvements that were achieved by the proposed approach, with a precision of Information Retrieval by Genetic Algorithm (GA-IR) for three datasets. Tables 4, 5, and 6 presents a comparison between our approach and GA-IR. Improvement average is calculated for three datasets and the results were 25.6666, 27.4444, and 45.2222 respectively.

Table 4. Comparison analysis of (K-means) - HPGA Approach and GA[26] in NPL Dataset(DS1)

| Recall | 0.1 | 0.2 | 0.3 | 0.4 | 0.5 | 0.6 | 0.7 | 0.8 | 0.9 | AVG |
|---|---|---|---|---|---|---|---|---|---|---|
| HPGA-(K-means) (p) | 0.9 | 0.87 | 0.84 | 0.77 | 0.74 | 0.66 | 0.58 | 0.46 | 0.38 | 0.6888 |
| GA-IR(P) | 0.88 | 0.66 | 0.59 | 0.44 | 0.4 | 0.31 | 0.27 | 0.19 | 0.15 | 0.4322 |
| Improvements % | 2 | 21 | 25 | 33 | 34 | 35 | 31 | 27 | 23 | 25.6666 |

Table 5. Comparison analysis of (K-means) - HPGA Approach and GA[26] in CISI Dataset(DS2)

| Recall | 0.1 | 0.2 | 0.3 | 0.4 | 0.5 | 0.6 | 0.7 | 0.8 | 0.9 | AVG |
|---|---|---|---|---|---|---|---|---|---|---|
| HPGA-(K-means) (p) | 0.89 | 0.84 | 0.78 | 0.76 | 0.69 | 0.55 | 0.51 | 0.47 | 0.44 | 0.6588 |
| GA-IR(P) | 0.8 | 0.55 | 0.48 | 0.39 | 0.36 | 0.28 | 0.24 | 0.2 | 0.16 | 0.3844 |
| Improvements % | 9 | 29 | 30 | 37 | 33 | 27 | 27 | 27 | 28 | 27.4444 |

Table (6) Comparison analysis of (K-means) - HPGA Approach and GA[26] in CACM dataset(DS3)

| Recall | 0.1 | 0.2 | 0.3 | 0.4 | 0.5 | 0.6 | 0.7 | 0.8 | 0.9 | AVG |
|---|---|---|---|---|---|---|---|---|---|---|
| HPGA-(K-means) (p) | 0.94 | 0.9 | 0.87 | 0.85 | 0.8 | 0.77 | 0.66 | 0.54 | 0.41 | 0.7488 |
| GA-IR(P) | 0.79 | 0.47 | 0.42 | 0.27 | 0.23 | 0.16 | 0.14 | 0.1 | 0.09 | 0.2966 |
| Improvements % | 15 | 43 | 45 | 58 | 57 | 61 | 52 | 44 | 32 | 45.2222 |

Finally, we compared the proposed approach with classic Information Retrieval (classic-IR) precision and the improvements were 34.4444% in NLP, 28.6666% in CISI, and 47% in CACM as shown in tables 7, 8 and 9.

Table (7) Comparison analysis of (K-means) - HPGA Approach and classic IR[20] in NPL Dataset(DS1)

| Recall | 0.1 | 0.2 | 0.3 | 0.4 | 0.5 | 0.6 | 0.7 | 0.8 | 0.9 | AVG |
|---|---|---|---|---|---|---|---|---|---|---|
| HPGA-(K-means) (p) | 0.9 | 0.87 | 0.84 | 0.77 | 0.74 | 0.66 | 0.58 | 0.46 | 0.38 | 0.6888 |
| Classic IR (P) | 0.73 | 0.5 | 0.44 | 0.34 | 0.31 | 0.24 | 0.22 | 0.17 | 0.15 | 0.3444 |
| Improvements % | 17 | 37 | 40 | 43 | 43 | 42 | 36 | 29 | 23 | 34.4444 |

Table 8. Comparison between (K-means) - HPGA Approach and classic IR[20] in CISI Dataset(DS2)

| Recall | 0.1 | 0.2 | 0.3 | 0.4 | 0.5 | 0.6 | 0.7 | 0.8 | 0.9 | AVG |
|---|---|---|---|---|---|---|---|---|---|---|
| HPGA-(K-means) (p) | 0.89 | 0.84 | 0.78 | 0.76 | 0.69 | 0.55 | 0.51 | 0.47 | 0.44 | 0.6588 |
| Classic IR (P) | 0.68 | 0.56 | 0.46 | 0.4 | 0.35 | 0.3 | 0.25 | 0.2 | 0.15 | 0.3722 |
| Improvements % | 21 | 28 | 32 | 36 | 34 | 25 | 26 | 27 | 29 | 28.6666 |

Table 9. Comparison analysis of (K-means) - HPGA Approach and classic IR[20] in CACM dataset(DS3)

| Recall | 0.1 | 0.2 | 0.3 | 0.4 | 0.5 | 0.6 | 0.7 | 0.8 | 0.9 | AVG |
|---|---|---|---|---|---|---|---|---|---|---|
| HPGA-(K-means) (p) | 0.94 | 0.9 | 0.87 | 0.85 | 0.8 | 0.77 | 0.66 | 0.54 | 0.41 | 0.7488 |
| Classic IR (P) | 0.72 | 0.45 | 0.37 | 0.25 | 0.22 | 0.16 | 0.14 | 0.11 | 0.09 | 0.2788 |
| Improvements % | 22 | 45 | 50 | 60 | 58 | 61 | 52 | 43 | 32 | 47 |

**6. CONCLUSIONS**

After the tests and research for this paper, we concluded an information retrieval performance improvement: (k-means) - HPGA achieved higher precision and better quality in document retrieval. Also a reduction of irrelevant results in user search was observed. Our results were determined by comparing three common datasets (NLP, CISI, and CACM) with Classic IR and GA. The range of precision improvements for three datasets with Classic-IR was [28% – 47%] while with GA-IR the precision was [25% - 45%].

**REFERENCES**

[1] C. D. Manning, P. Ragahvan, and H. Schutze, "An Introduction to Information Retrieval", no. c. 2009.
[2] J. M. Kassim and M. Rahmany, "Introduction to Semantic Search Engine," *2009 Int. Conf. Electr. Eng. Informatics*, vol. 02, no. August, pp. 380–386, 2009
[3] S. M. Weiss, N. Indurkhya, T. Zhang, and F. J. Damerau, "Information Retrieval and Text Mining," *Springer Berlin Heidelb., no. Fundamentals of Predictive Text Mining*, pp. 75–90, 2010.
[4] Y. Wang, "Design of Information Retrieval System Using Rough Fuzzy Set," *TELKOMNIKA Indones. J. Electr. Eng.*, 2014.
[5] Y. Djenouri and *et al*," Fast and Effective Cluster-based Information Retrieval using Frequent Closed Itemsets", *InformationSciences,* 2018.
[6] C. Cobos and *et al*, "Web document clustering based on Global-Best Harmony Search, K-means, Frequent Term Sets and Bayesian Information Criterion", *IEEE*, 2010.
[7] I. Irmawati, "Information Retrieval in Documents using Vector Space Model," *J. Ilm. FIFO*, 2017
[8] R.F.Hassan and *et al*, " Improving the Web Indexing Quality through A Website-Search Engine Coactions", *International Journal of Computer and Information Technology* (ISSN: 2279 – 0764) Volume 03, Issue 02, March 2014.
[9] S. S. Tandel, A. Jamadar, and S. Dudugu, "A Survey on Text Mining Techniques," *5th Int. Conf. Adv. Comput. Commun. Syst* pp. 1022–1026, 2019.
[10] S.H.Toman and *et al,* " Content-Based Audio Retrieval by using Elitism GA-KNN Approach", *Journal*

*of AL-Qadisiyah for computer science and mathematics*, Vol.9,No.1 ,2017.
[11] A. Konar, "Artificial intelligence and soft computing: behavioral and cognitive modeling of the human brain", *Jadavpur University, CRC Press LLC*, 2000.
[12] T. Munakata, "Fundamentals of the New Artificial Intelligence", Second Edition, *Springer*,2008.
[13] C. C. Aggarwal and C. Xhai, "A survey of text clustering algorithms," in *Mining text data*, 2012.
[14] A.M. Siregar and A. Puspabhuana, "Improvement of term weight result in the information retrieval systems," *in Proceedings of 4th International Conference on New Media Studies*, 2017.
[15] T. Tokunaga, T. Tokunaga, I. Makoto, and I. Makoto, "Text categorization based on weighted inverse document frequency," *Spec. Interes. Groups Inf. Process Soc. Japan* (SIG-IPSJ, 1994.
[16] Ph. Simon and S. S. Sathya, "Genetic Algorithm for Information Retrieval", *IEEE*, 2009.
[17] Z. Wang and B. Feng, "Optimal Genetic Query Algorithm for Information Retrieval", *Springer*, 2009.
[18] H. Imran, "Genetic Algorithm Based Model for Effective Document Retrieval", *Springer*, 2011.
[19] P. Pathak, M. Gordon and W. Fan, "Effective Information Retrieval using Genetic Algorithms based Matching," *Hawaii International Conference on System Sciences*, IEEE, 2000.
[20] M. Ebrahimi and A. Jahangirian, "A hierarchical parallel strategy for aerodynamic shape optimization with genetic algorithm", *Scientia Iranica*,22(6), 2379-2388, 2015
[21] Kh. I. Abuzanouneh, "Parallel and Distributed Genetic Algorithm with Multiple-Objectives to Improve and Develop of Evolutionary Algorithm", *International Journal of Advanced Computer Science and Applications*, Vol. 7, No. 5, 2016.
[22] J. Marini, "The Document Object Model, Processing Structured Documents", *McGraw-Hill/Osborne*, 2002.
[23] S-H. Cha, "Comprehensive Survey on Distance/Similarity Measures between Probability Density Functions", *International Journal of Mathematical Models and Methods in Applied Sciences*, Issue 4, Volume 1, 2007.
[24] L. T. Su, "The relevance of recall and precision in user evaluation," J. Am. Soc. Inf. Sci., 1994
[25] M. Junker, R. Hoch, and A. Dengel, "On the evaluation of document analysis components by recall, precision, and accuracy," *in Proceedings of the International Conference on Document Analysis and Recognition, ICDAR*, 1999.
[26] A. A. Radwan and et al, "Using Genetic Algorithm to Improve Information Retrieval Systems", *World Academy of Science, Engineering and Technology International Journal of Computer and Information Engineering* Vol:2, No:5, 2008.

**BIOGRAPHIES OF AUTHORS**

| | |
|---|---|
| **Sarah Hussein Toman** | Lecturer in Roads and Transport department , Iam working as an academic staff at Al-Qadisiyah University, College of Engineering , my major research focusing on  web technology and Information Retrieval. B.Sc ,M.Sc, in Computer Science. |
| **Mohammed Hamzah Abed** | He recevied the B.Sc degree in Computer Science from University of Al-Qadisiyah , Iraq 2008, M.Sc degree in Computer Science from B.A.M. University ,india 2011.currently, he works as a lecturer at the Department of Computer Science, University of Al-Qadisiyah and he doing the researches in the Medical Image Processing and machine learning. |
| **Zinah Hussein Toman** | Lecturer in Computer Science , She is working as an academic staff at Al-Qadisiyah University, College of Engineering , my major research focusing on  web technology and IoT. B.Sc ,M.Sc, in Computer Science. |